\documentclass[sigconf]{acmart}
\usepackage[scr=boondox,cal=esstix]{mathalpha}
\usepackage{array}
\usepackage{amsmath}
\usepackage[linesnumbered,ruled,vlined]{algorithm2e}
\AtBeginDocument{%
  }

\setcopyright{acmlicensed}
\copyrightyear{2018}
\acmYear{2024}
\acmDOI{XXXXXXX.XXXXXXX}

\acmConference[0000]{The 0000 Conference on 0000}{0000. 00--00, 0000}{0000, 0000, 0000}
\acmISBN{978-1-4503-XXXX-X/18/06}

\newcommand{\thickhline}{%
    \noalign {\ifnum 0=`}\fi \hrule height 1pt
    \futurelet \reserved
}
\begin{document}

\title{Rethinking Radiology Report Generation via Causal Inspired Counterfactual Augmentation}

\author{Xiao Song}
\email{x.song.matt@gmail.com}
\orcid{0000-0001-6352-6542}
\affiliation{%
  \institution{Nanjing University}
  \city{Suzhou}
  \state{Jiangsu}
  \country{CHN}
}


\author{Jiafan Liu}
\email{jiagonglou21@gmail.com}
\affiliation{%
  \institution{South China Agricultural University}
  \city{Guangzhou}
  \state{Guangdong}
  \country{CHN}}

\author{Yan Liu}
\email{liuyan23@nankai.edu.cn}
\affiliation{%
 \institution{Nankai University}
 \city{Tianjin}
 \state{Hebei}
 \country{CHN}}

\author{Yun Li}
\email{liyun76@mail.sysu.edu.cn}
\author{Wenbin Lei}
\affiliation{%
  \institution{The First Affiliated Hospital, Sun Yat-sen University}
  \city{Guangzhou}
  \state{Guangdong}
  \country{CHN}}

\author{Ruxin Wang}
\email{rx.wang@siat.ac.cn}
\authornotemark[1]
\affiliation{%
 \institution{Shenzhen Institute of Advanced Technology, Chinese Academy of Sciences}
 \city{Shenzhen}
 \state{Guangdong}
 \country{CHN}}

\renewcommand{\shortauthors}{Xiao Song, et al.}

\begin{abstract}
Radiology Report Generation (RRG) draws attention as a vision-and-language interaction of biomedical fields. Previous works inherited the ideology of traditional language generation tasks, aiming to generate paragraphs with high readability as reports. 
Despite significant progress, the independence between diseases—a specific property of RRG—was neglected, yielding the models being confused by the co-occurrence of diseases brought on by the biased data distribution, thus generating inaccurate reports.
In this paper, to rethink this issue, we first model the causal effects between the variables from a causal perspective, through which we prove that the co-occurrence relationships between diseases on the biased distribution function as confounders, confusing the accuracy through two backdoor paths, i.e. the  Joint Vision Coupling and the  Conditional Sequential Coupling.
Then, we proposed a novel model-agnostic counterfactual augmentation method that contains two strategies, i.e. the Prototype-based Counterfactual Sample Synthesis (P-CSS) and the Magic-Cube-like Counterfactual Report Reconstruction (Cube), to intervene the backdoor paths, thus enhancing the accuracy and generalization of RRG models. 
Experimental results on the widely used MIMIC-CXR dataset demonstrate the effectiveness of our proposed method. Additionally, a generalization performance is evaluated on IU X-Ray dataset, which verifies our work can effectively reduce the impact of co-occurrences caused by different distributions on the results. 
\end{abstract}




\maketitle

\section{Introduction}
\label{sec:intro}

Radiology Report Generation (RRG), which automatically generates free-text reports to describe the visual findings of radiographs, has attracted increasing attention recently in the interaction between vision and language in health informatics.
Specifically, RRG generates reports in the form of free-text paragraphs after first identifying the visual information from radiographs.
The generated reports contain a series of sentences, each of which independently describes certain types of normal or abnormal anatomic structures~\cite{tanida2023interactive, hou2023organ}.

Previous methods~\cite{Li_Liang_Hu_Xing_2019, Zhang_Wang_Xu_Yu_Yuille_Xu_2020, Liu_2021_CVPR, Wang2022, 10016250} followed the ideology of traditional machine translation and image-to-text generation tasks like Image Captioning~\cite{Vinyals_2015_CVPR, Karpathy_2015_CVPR}, aiming to generate a paragraph with high natural language generation metrics. 
Many of these cutting-edge research works \cite{Li_Liang_Hu_Xing_2019, Zhang_Wang_Xu_Yu_Yuille_Xu_2020, Liu_2021_CVPR, Wang2022} integrated the disease co-occurrence relationships that are summarized from the data distribution of the training sets as prior knowledge into RRG models.
Most of these recent methods utilized the co-occurrence to enhance the dependencies between diseases to force models to fit the distribution of training data and improve the natural language generation performance.
While these research efforts have achieved significant progress, the integration of disease co-occurrence implicitly restrains the models' ability to distinguish diseases independently. The model may have abandoned identifying specific disease features and relied solely on one disease to infer another, which confuses the recognition and generation of visual reality.

\begin{figure*}
\centering
\includegraphics[width=.66\linewidth]{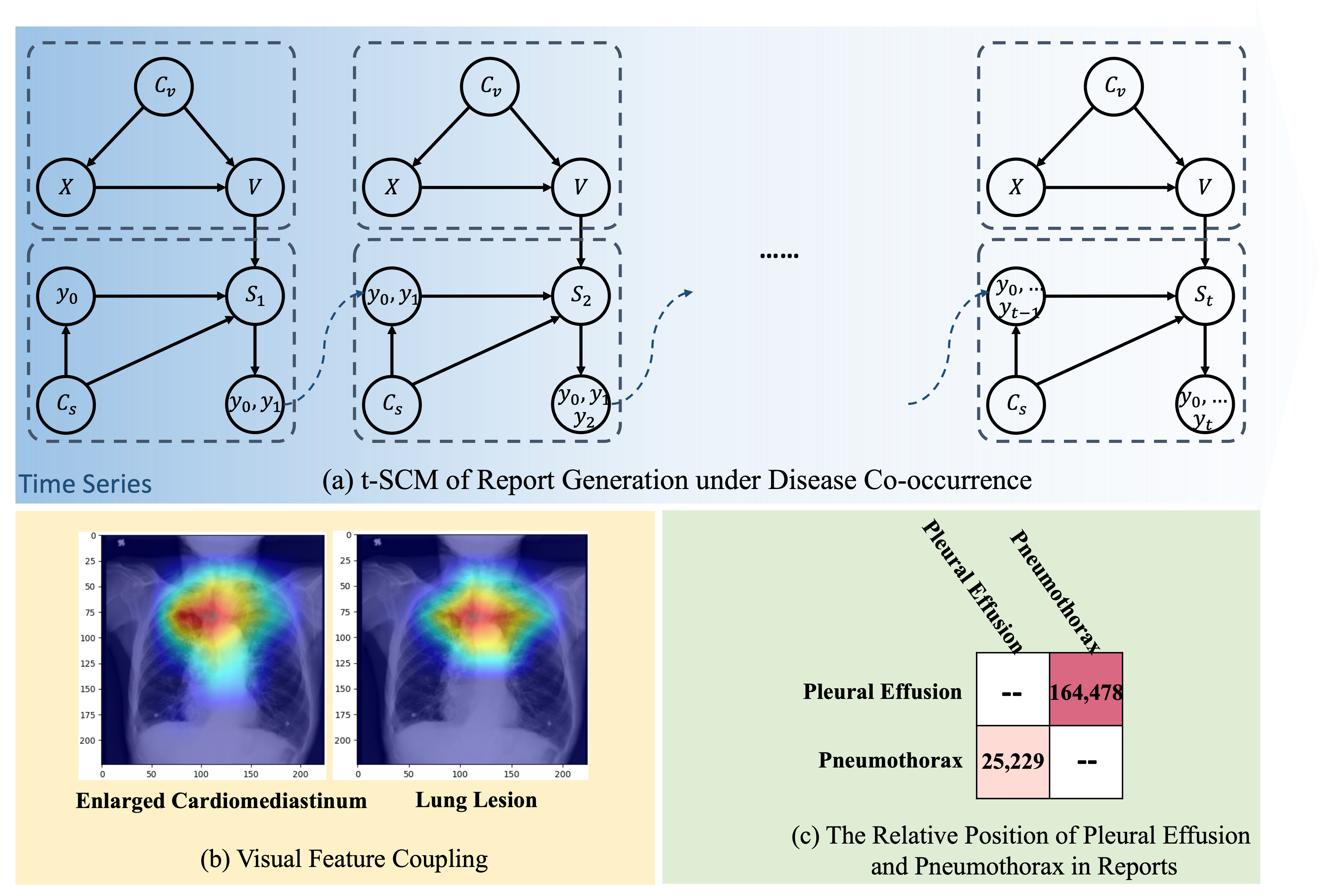}
\caption{\label{fig:causalgraph}{(a) is the SCM graph under the Joint Vision Coupling $C_v$ and Conditional Sequential Coupling $C_s$. (b) and (c) are two examples with respect to the joint vision coupling and conditional sequential coupling.
}}
\end{figure*}

However, to our knowledge, there is no existing work that has explored this issue. 
Therefore, in this paper, to rethink this problem thoroughly, we reason about the causes and effects of the disease co-occurrence relationship in the field of RRG from a novel perspective of statistics and causality. 
Specifically, we first find the cause of disease co-occurrence and discuss whether it is beneficial to RRG from a novel perspective of causality. Then, the two aspects of effects shown in Fig.\ref{fig:causalgraph} are discovered with structure causal models, where the Joint Vision Coupling and the Conditional Sequential Coupling implicitly decrease the accuracy of reports. 

Based on the rethinking and reasoning of co-occurrences in RRG, we propose a simple and effective model-agnostic counterfactual augmentation method to solve the problems, which contains two strategies, i.e., the Prototype-based Counterfactual Sample Synthesis (P-CSS) and the Magic-Cube-like Counterfactual Report Reconstruction (Cube). 
Specifically, the P-CSS strategy, which randomly masks visual features and their corresponding sentences in the recognition process, is proposed to address the visual coupling and improve independence when generating the sentence of a certain visual finding. 
Additionally, the Cube strategy that randomly reconstructs the sentence sequences is proposed to prevent models from being disturbed by the previously generated sentence when generating a new one. 
Experimental results and further analyses on two widely used datasets, MIMIC-CXR~\cite{johnson2019mimic} and IU X-Ray~\cite{demner2016preparing}, demonstrate the capability of both sub-methods to improve the accuracy of generated reports and achieve better performance when combined together. Overall, our contributions are as follows:
\begin{itemize}

\item To our knowledge, this is the first work to discuss the disease co-occurrence in the RRG task from the causal view and explore the effects of the two spurious confounders, i.e., the Joint Vision Coupling $C_v$ and the Conditional Sequential Coupling $C_s$, on RRG models from a novel perspective of statistics and causality. We also propose a novel temporal structure causal model to model the causal effect of $C_v$ and $C_s$ during the process of report generation.

\item We propose a simple and effective model-agnostic counterfactual augmentation method that contains the Counterfactual Sample Synthesis strategy and the Counterfactual Report Reconstruction strategy to solve the problems caused by the two spurious coupling confounders respectively. 

\item Experiments are conducted on the widely used MIMIC-CXR dataset and a generalization study is implemented on IU X-Ray dataset, which further demonstrates the value of our rethinking on RRG task and the effectiveness of our proposed method.
\end{itemize}

\section{Related Works}

\subsection{Radiology Report Generation}
Radiology Report Generation (RRG) is an image-to-text generation task akin to Image Captioning~\cite{Vinyals_2015_CVPR, Karpathy_2015_CVPR}. 
In contrast to Image Captioning, which creates a brief sentence expressing key information in natural images, the labels of RRG are free-text long paragraphs that detail the multi-disease findings.~\cite{demner2016preparing, johnson2019mimic}. 
Therefore, the previous works proposed to use HLSTM~\cite{krause2017hierarchical} that hierarchically generate the topics and sentences with two LSTM layers~\cite{jing_2018, NEURIPS2018_e0741335, zhang2023weakly, shi2023granularity, liu2024multi}, or employed the long-term memory ability of Transformer~\cite{vaswani2017attention} model to directly produce a long paragraph~\cite{chen2020generating, li2023dynamic, tanida2023interactive, hou2023organ, hou2023recap}. 
Specifically, 
Zhang et al.~\cite{zhang2023weakly} and Shi et al.~\cite{shi2023granularity} conducted the cross-modal alignment of brain CT images and sentences, then generated reports with a two-layer hierarchical LSTM. 
Chen et al.~\cite{chen2020generating} designed a memory driven conditional layer normalization module to improve the disease semantic memory of Transformer. 

To generate reports with higher natural language generation metrics, many advanced works~\cite{Li_Liang_Hu_Xing_2019, Zhang_Wang_Xu_Yu_Yuille_Xu_2020, Liu_2021_CVPR, Wang2022, doushixin} proposed to integrate disease co-occurrence mined from training data distribution. 
For example, Li et al. \cite{Li_Liang_Hu_Xing_2019} proposed a knowledge-driven encode, retrieve, and paraphrase approach that learns to enhance the relations between anatomic terminologies with Graph Transformer. 
Zhang et al. \cite{Zhang_Wang_Xu_Yu_Yuille_Xu_2020} assumed that abnormalities on the same body part might have a strong correlation with each other and share many features, thus they proposed a knowledge graph that groups chest diseases occurring in the same organ or tissue for assisting radiology report generation.
Liu et al. \cite{Liu_2021_CVPR} proposed a prior knowledge explorer to find the most relevant chest terminologies from that knowledge graph \cite{Zhang_Wang_Xu_Yu_Yuille_Xu_2020} and to match the language from the existing report corpus, which improves the report consistency. 
Wang et al. \cite{Wang2022} enforced the co-occurrence of chest terminologies to be incorporated into the final reports.
These methods improved RRG model performance, particularly in paragraph coherence, which promoted the injection of disease co-occurrence as a common strategy in RRG models.


\subsection{Counterfactual Augmentation for vision-and-language grounding problems}
Counterfactual augmentation (CA), which produces augmented data by making counterfactual modifications to a subset of causal factors~\cite{pitis2020counterfactual}, has recently emerged as a method to mitigate confounding biases in the training data for a machine learning model~\cite{liu2021counterfactual}.
Additionally, CA is an important model-agnostic causal intervention method to prevent the models from being disturbed by confounders, and many researchers have explored it in various tasks. 
For example, in the interaction of vision and language, some works~\cite{agarwal2020towards,chen2023counterfactual} try to synthesize counterfactual samples for VQA. In the vision-only problems, Yao et al.~\cite{xiao2023masked} manufactured counterfactual samples through masking and refilling the patches of images to robustly fine-tune the pre-trained vision models. In the language-only problems, to improve the robustness of neural answer selection models, Zhong et al.~\cite{zhong2022reducing} intervened in the spurious correlations between prediction labels and input features by removing the feature dependencies and language biases in answer selection. However, in contrast to natural images that are much easier and more mature to precisely process with methods like Object Detection or Semantic Segmentation, radiographs containing dozens of diseases are hard to accurately detect or segment. Besides, the anatomic structures of certain types of diseases frequently overlap in space, thus it is challenging and unrealizable to plot out each disease independently. Therefore, the counterfactual augmentation methods that have made great achievements in natural image and language processing can not be applied directly to the RRG task. 


\begin{table}
\centering \small
\caption{Data distribution bias of \textit{Pleural Effusion} ($\mathcal{B}$) conditioned on \textit{Pneumothorax} ($\mathcal{A}$), i.e. $P(\mathcal{B}^*|\mathcal{A}^*)$, in MIMIC-CXR datasets. $*$ contains `+' and `-', denoting abnormal and normal respectively. The integer numbers are the amount of data items and the decimal number in $(\text{-})$ are the probabilities of $\mathcal{B}^*$ when given $\mathcal{A}^*$.} 
\label{table:simpson}
\makebox[\linewidth][c]{
\begin{tabular}{lcc}
\toprule[1pt]
  & $\mathcal{A}^{+}$ & $\mathcal{A}^{-}$ \\
  & 7,667 & 195,425
\tabularnewline
\midrule[0.5pt]
$\mathcal{B}^{+}$ & 3,552 (0.463) & 30,988 (0.159) \\
$\mathcal{B}^{-}$ & 1,345 (0.175) & 153,822 (0.787) \\
\bottomrule[1pt]
\end{tabular}}
\end{table}

\section{Rethinking Co-occurrence from A Causal View}
\label{sec:rethinking}
Disease co-occurrence, denoted as $\mathcal{dc}$, has been widely used in previous works~\cite{Li_Liang_Hu_Xing_2019, Liu_2021_CVPR, Wang2022}. The model can more ``conveniently'' generate corresponding text descriptions by utilizing the co-occurrence relationship in the training samples between different diseases as a shortcut to improve the performance of Radiology Report Generation (RRG) models, especially on the Natural Language Generation (NLG) metrics. 
However, in medical scenarios, accuracy for diagnosis is unquestionably more crucial compared to the NLG. The spurious or biased co-occurrence relationships may lead to model misidentification. Thus, it is important to rethink the role of $\mathcal{dc}$ in RRG.

\subsection{Dose $\mathcal{dc}$ always denote a causal relationship for RRG?}

In RRG, $\mathcal{dc}$ is widely used to improve the NLG metrics. However, some co-occurrences existing in the training set may mislead the recognition of models.
To simplify the elucidation, we select two types of diseases that are always seen as a couple of $\mathcal{dc}$, i.e. the \textit{Pleural Effusion} denoted as $\mathcal{A}$ and the \textit{Pneumothorax} denoted as $\mathcal{B}$. We analyze the data distributions of $\mathcal{B}$ when conditioned on $\mathcal{A}$ denoted as $P(\mathcal{B}^*|\mathcal{A}^*)$, as shown in Tab.~\ref{table:simpson}, in which $^*$ denotes the subsets where the disease is positive $^+$ or negative $^-$. From Tab. \ref{table:simpson}, it can be observed that when the patient has pleural effusion (given $\mathcal{A}^+$), the probability of simultaneously suffering from pneumothorax is much higher than not having pneumothorax ($P(\mathcal{B}^+|\mathcal{A}^+) > P(\mathcal{B}^-|\mathcal{A}^+)$), which suggests the pleural effusion may cause pneumothorax. However, from a medical perspective, there is no causal relationship between pleural effusion and pneumothorax. Hence, utilizing this spurious co-occurrence would result in a wrong effect. Similarly, in this data distribution, given $\mathcal{A}^-$, the condition distribution $P(\mathcal{B}^-|\mathcal{A}^-)$ occupies a dominant position. Employing this biased co-occurrence also provides a shortcut for generating a sequence description with categories $\mathcal{A}^-$ and $\mathcal{B}^-$. The above examples demonstrate the two most common ``harmful'' co-occurrence relationships caused by sampling bias. It implies that we should weaken the consideration of the contribution of this kind of co-occurrence to natural language indicators and instead break down false or biased co-occurrence relationships, focusing more on the accuracy of actual disease identification.

\subsection{What effects does $\mathcal{dc}$ have on RRG?}
To further answer the question ``What effects does $\mathcal{dc}$ have on RRG?'', we employ the Structure Causal Model (SCM) to describe the relationship of relevant variables and how they interact with each other, which could be represented as the directed acyclic graphs $G=\{V, E\}$ with the set of variables $V$ and the causal-effect relationships $E$. 
Fig.\ref{fig:causalgraph} (a) shows the t-SCM which contains six variables $X$, $V$, $S_t$, $C_v$, $C_s$, and $Y=\{y_1,...,y_t,...,y_T\}$, respectively. $X$ denotes the input radiograph, and $V$ refers to the visual representation learned from the image. $S_t$ indicates the $t$th linguistic features integrating the visual representation $V$ and generated sequence sentence information $\{y_0, y_1,...,y_{t-1}\}$, where $y_0$ is a start token and $y_t$ is the $t$th generated sentence. 
In this paper, we mainly investigate two confounder effects for the model learning, i.e. the Joint Vision Coupling $C_v$ and the Conditional Sequential Coupling $C_s$. To eliminate the spurious correlation by the $C_v$ and $C_s$. We introduce the backdoor adjustment method with the \emph{do}-calculus to block the effect of these hidden confounders $C=\{C_v,C_s\}$:
\begin{equation} 
    \begin{aligned}
        P(y_t|do(V, S_t)) &= \sum_{c \in C} P(y_t|V, S_t, c)P(c) \\ &=\sum_{c_v \in C_v} P(y_t|V, c_v)P(c_v) + \sum_{c_s \in C_s} P(y_t|S_t, c_s)P(c_s).
    \end{aligned}
    \label{cut_off_doXc}
\end{equation}
Specifically, we analyse these two confunders as follows: 
\subsubsection{Effects of Joint Vision Coupling}~\\
\label{label:cj}
Due to the spatial overlap of the visual positions of many different types of diseases in 2D radiographic images, RRG models are hard to independently identify each disease based on the features that truly matter to its category. Therefore, as shown in Fig.\ref{fig:causalgraph} (a), we define this joint vision coupling as $C_v$. $X \rightarrow V$ represents the direct effect of input radiograph $X$ on visual features $V$. $C_v \rightarrow X$ is the visual features extracted by the models when conditioned on the joint visual coupling. The ``back door'' path $C_v \leftarrow X \rightarrow V \rightarrow ... \rightarrow Y$ bridges the spurious path between $C_v$ and $Y$, which may result in a wrong prediction based on the confounder $C_v$. Fig.\ref{fig:causalgraph} (b) shows an example that two unrelated diseases overlap in some areas in 2D visual representation.
To break the effect of visual co-occurrence, we employ the decoupled representation of different diseases and then randomly mask one decoupled visual features related to a specific disease (in practice, the corresponding text description would also be deleted). By the above method, a simple and effective approximate estimate of $P(y_t|V, c_v)$ can be obtained as follows: 
\begin{equation}
    \begin{aligned}
       \sum_{c_v \in C_v} P(y_t|V, c_v)P(c_v) \approx \sum_{n=1}^{K} \sum_{\hat{c}_v \in \hat{C}^n_v} P(y_t|V, \hat{c}_v)P(\hat{c}_v), 
    \end{aligned}
    \label{ori_data_augmentation}
\end{equation}
where $K$ indicates the category numbers. $\hat{C}^n_v$ is a random mask strategy of category $n$. One available implementation approach can be found in Sec. 4.2.




\subsubsection{Effects of Conditional Sequential Coupling} ~\\
\label{label:cc}
In auto-regressive problems, the prediction of the next token is conditioned on the previous outputs, which means that the previous output will affect the results of subsequent outputs. Due to some sequential bias in the dataset, when the describing sentence of categories $\mathcal{A}$, denoted as $S_\mathcal{A}$, frequently appears in front of the describing sentence of categories $\mathcal{B}$, denoted as $S_\mathcal{B}$, the models will learn to establish the conditional coupling of $\mathcal{A}$ and $\mathcal{B}$ (even if $\mathcal{A}$ and $\mathcal{B}$ are independent). Then the generated $S_\mathcal{B}$ based on $p(S_\mathcal{B})$ is disturbed by the conditional probability $p(S_\mathcal{B} | S_\mathcal{A})$. Specifically, in the RRG task, the sentences in the report are almost irrelevant, which means that the sequence of sentences has no impact on the accuracy of the report. For example, \textit{``The lungs are clear. No pleural effusions.''} and \textit{``No pleural effusions. The lungs are clear.''} are exactly the same. However, the frequency of some combinations shows overwhelming numbers as shown in Fig.\ref{fig:causalgraph} (c). In this paper, we define this phenomenon as the Conditional Sequential Coupling $C_s$. As illustrated in Fig.\ref{fig:causalgraph} (a), $V \rightarrow S_t$ represents the direct effects of the visual representation $V$ to linguistic features $S_t$. $M \rightarrow \{y_0,\dots,y_{t}\}$ is the direct effect of the linguistic features on the generation of the $t$th sentence of the report. Similarly, $\{y_0,...,y_{t-1}\} \leftarrow C_s \rightarrow S_t \rightarrow \{y_0,\dots,y_{t}\}$ conducts another backdoor path. Accordingly, the backdoor path causes the models to learn to generate the next sentence influenced by $C_s$, which reduces the concentration on visual reality. Additionally, because of the seriously biased sentence sequence of reports in RRG, $\mathcal{dc}$ implicitly manifests as a co-occurrence relationship in the form of sequences. Therefore, the accuracy of the generated reports is implicitly damaged because of this biased sequential relationship.




According to Eq. \ref{cut_off_doXc}, since enumerating all confounders $c_s \in C_s$ is impractical, we employ a random reordering strategy to estimate $P(y_t|S_t, c_s)$ in training stage:  
\begin{equation}
    \begin{aligned}
       \sum_{c_s \in C_s} P(y_t|S_t, c_s)P(c_x) \approx \sum_{\hat{c}_s \in \hat{C}_s} P(y_t|S_t, \hat{c}_s)P(\hat{c}_s) 
    \end{aligned},
    \label{ori_data_augmentation}
\end{equation}
where $\hat{C}_s$ represents the set of strategies for sorting sentences $\{y_1,...,y_{t-1}\}$ and $c_s$ indicates one specific order.

\begin{figure*}
\centering
\includegraphics[width=.85\linewidth]{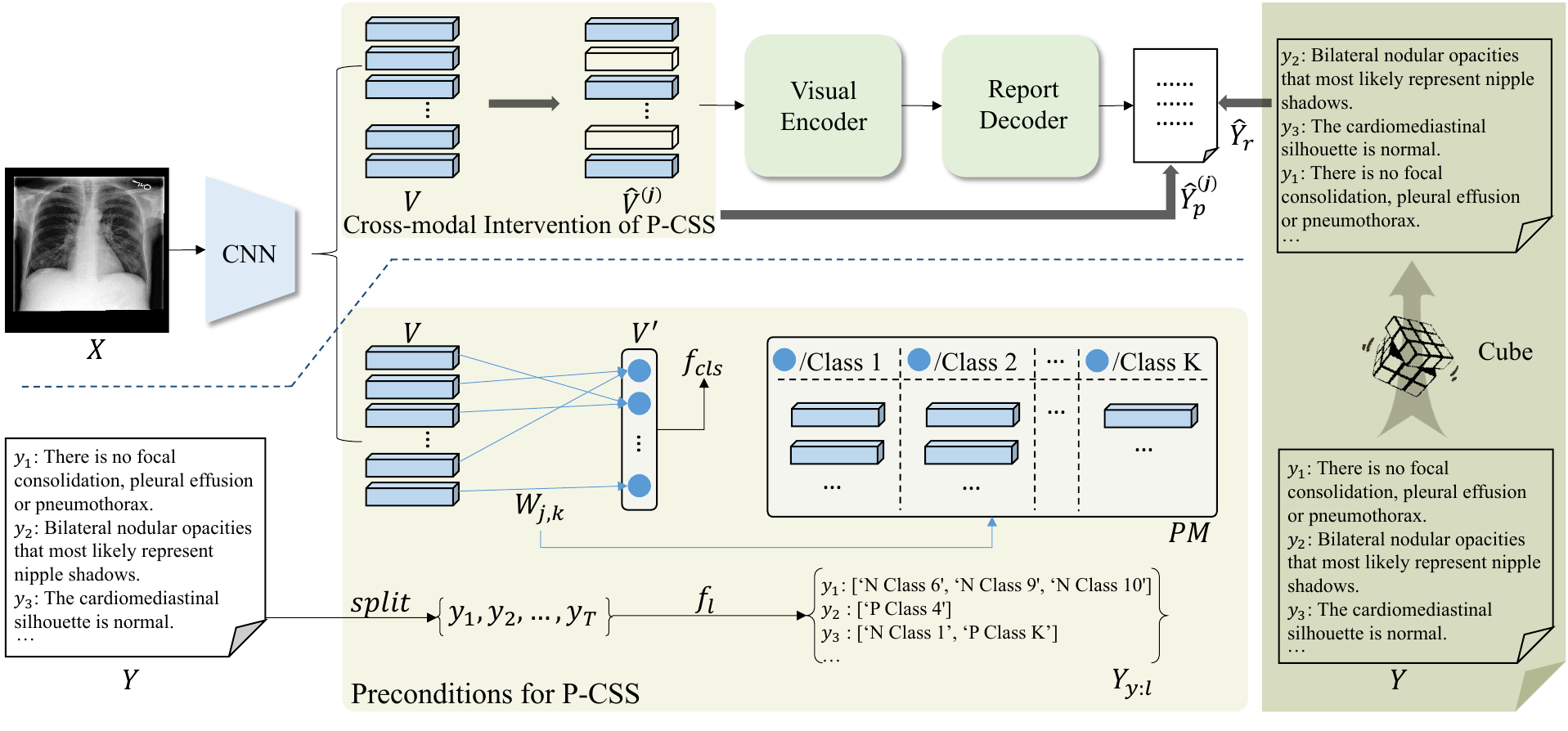}
\caption{\label{fig:method}{Overview of our proposed counterfactual augmentation method, which contains (a) the Prototype-based Counterfactual Sample Synthesis (P-CSS) and (b) the Magic-Cube-like Counterfactual Report Reconstruction(Cube). (a) first prepares the Class-wise Visual Prototype Matrix $PM$ and the Sentence-level Class Labeling $Y_{y:l}$, then conducts counterfactual cross-modal intervention on the training set. (b) randomly disturbs the sequence of sentences in the report.
}}
\end{figure*}

\section{Methodology}
We design two counterfactual strategies, i.e. the Prototype-based Counterfactual Sample Synthesis (P-CSS) and the Magic-Cube-like Counterfactual Report Reconstruction (Cube), to respectively solve the problems caused by the aforementioned Joint Visual Coupling $C_v$ and Conditional Sequential Coupling $C_s$. In this section, we will first model the Radiology Report Generation task as a mathematical model, then introduce the proposed P-CSS and Cube strategies in detail.

\subsection{Modeling Radiology Report Generation}
\label{lebel:modeling}
Given a radiograph $X$, the goal is to generate a long medical report $\tilde{Y}=\{\tilde{y_1}, \tilde{y_2},...,\tilde{y_T}\}$ with $T$ sentences that describes the information of diseases shown in the radiograph. 
Thus, the Radiology Report Generation task aims to learn the parameters $\theta$ of model by maximizing the probability of the correct description:
\begin{eqnarray}
\theta^{*}=arg \; max \log  p(Y|X; \theta), \label{eqn-bg1}
\end{eqnarray}
where $Y = \{y_1, y_2,...,y_T\}$ is the ground-truth report.
Thus, it is common to apply the chain rule to model the joint probability over $\{y_1, y_2,...,y_T\}$:
\begin{eqnarray}
log\,p(Y|X)=\sum{log\,p(y_t|X,y_{1:t-1})}. \label{eqn-bg2}
\end{eqnarray}

\subsection{Prototype-based Counterfactual Sample Synthesis}
Based on the preceding analysis in Sec. \ref{label:cj}, we propose a Prototype-based Counterfactual Sample Synthesis (P-CSS) strategy to intervene the effect of $C_v$ in RRG, hence preserving the models' capacity to identify the exclusive visual traits of each category. The overall workflow of P-CSS is shown in Fig.~\ref{fig:method}. P-CSS first establishes the class-wise visual prototype matrix $PM$ based on the class activation mapping to the spatial features. Each column in $PM$ indicates a group of class-activated spatial positions in the CNN extracted visual features. 
Simultaneously, P-CSS splits the reports into the form of sentence list then obtains the label of each sentence $L_{y_t}$ with automatic labeler.
Finally, based on $PM$ and $L_{y_t}$, P-CSS synthesizes class-wise counterfactual samples to break the effects of $C_v$ on RRG from the aspect of data augmentation.
In the remaining portion of this subsection, we will detail the establishing of $PM$, the Sentence-level Class Labeling, and the Counterfactual Cross-modal Intervention in order.
~\\
\subsubsection{Preconditions for P-CSS}~\\
\textbf{Class-wise Visual Prototype Matrix.}
$PM$ memorizes the activated spatial positions for each category prototype. 
Specifically, given a radiology $X$, we extract its visual feature map $V\in \mathbb{R}^{H\times W\times D}$ using CNN-based models pre-trained on the multi-label classification function $f_{cls}$ as shown in Fig.~\ref{fig:method}
and flatten $H$ and $W$ as $V\in \mathbb{R}^{(H W)\times D}$.
Then, we project the visual representation $V$ into the feature map of $K$ categories as follows:
\begin{eqnarray}
v^{'}_j = \sum_{k=1}^{HW} W_{j,k}\cdot v_k, \quad j \in\{1,...,K\}, 
\end{eqnarray}
where $K$ is the number of disease categories, $v^{'}_j$ is the spatial features that are related to the $j$th category,
and $v_k$ is the feature of $k$th spatial position in $V$.
$W_{j,k}$ is the weights of projection function that reflects the importance of $k$ to class $j$.
Therefore, we can establish the class-wise vision prototype based on $W_{j,k}$, where a threshold $\tau$ is set to filter the positions which is strongly relative to each category. This operation can be formulated as:
\begin{eqnarray}
PM_{j} = \begin{cases}
1 \text{,}& W_{j,k} \textgreater \tau\\
0 \text{,}& \text{else}
\end{cases},
\end{eqnarray}
where the $PM_{j} \in \mathbb{R}^{1 \times HW}$ is the visual prototype of $j$th category that indicates the strong related positions in $X$.~\\

\textbf{Sentence-level Class Labeling.}
$PM$ links the visual features and the sentences of each category. To align different visual features with their corresponding textual expressions, category information is required for each sentence in reports, which is, however, not provided in the datasets. 
To address this problem, we use CheXbert~\cite{smit2020combining}, an automatic chest report labeler, to label each sentence.
Specifically, given the raw reports $Y$, we first split it into the list of sentences, denoted as $\{y_1,...,y_T\} = split(Y)$. Each item $y_t$ is a sentence that contains a series of tokens. 
Then, we respectively label each item as:
\begin{eqnarray}
L_{y_t} = f_l(y_t),
\end{eqnarray}
where $f_l$ is the automatic labeler and $L_{y_t}$ is the labels of sentence $y_t$. 
Formally, to simplify further operations, we design the presentation of each category label as \textit{``P/N/U CategoryName''}, where \textit{P/N/U} are \textit{Positive/Negative/Uncertain} respectively. 
For example, if the \textit{negative Consolidation} is identified in $y_t$, then $L_{y_t}$ can be presented as \textit{"N Consolidation"}. 
Additionally, because some samples might describe several categories of diseases in one sentence, $L_{y_t}$ is set to be the list of labels that contains all the described categories.
Finally, the labeled sentences could be represented as a dict $Y_{y:l}=\{y_t : L_{y_t}\}_{t=1}^{T}$.
~\\



\begin{algorithm} 
\caption{Counterfactual Cross-modal Intervention}\label{alg1}
  \KwIn{the original Training Sample $(V, Y)$, the class-wise visual prototype matrix $PM$, the sentence-level class labels $Y_{y:l}$.}
  \KwOut{intervened sample $(\hat{V}^{(j)}, \hat{Y}^{(j)}_p)$.}  
  \SetKwFunction{FMain}{Main}
  \SetKwProg{Fn}{Function}{:}{}
  \Fn{\FMain{V, Y}}{
  \SetKwFunction{FMain}{InterventionOnVision}
  \SetKwProg{Fn}{Function}{:}{}
  \Fn{\FMain{V, j}}{
$PM_j  \leftarrow j$ th prototype from $PM$\\
$\hat{V}^{(j)} = V \; \odot \;(1- PM_{j}^{T})$ \\
\KwRet $\hat{V}^{(j)}$\\
  }
random class $j$\\
temp = [j] \\
\If{$j$ is described in $Y$ }
{temp.extend($l_j$) \\
pop($y_j$)\\
$\hat{Y}^{(j)}_p = concate([y_t]_{t=1}^{T^{'}})$
}
$\hat{V}^{(i)} = V$ \\
\For{$i$ in temp}{
$\hat{V}^{(i)}$ = InterventionOnVision($\hat{V}^{(i)}$, $i$)
}
  \KwRet $(\hat{V}^{(j)}, \hat{Y}^{(j)}_p)$
}
\end{algorithm}

\subsubsection{Counterfactual Cross-modal Intervention}~\\
After preparing the class-wise visual prototype matrix $PM$ and the labeled sentences dict $Y_{y:l}$, P-CSS conducts the prototype-based counterfactual sample synthesis to intervene the effects of Joined Visual Coupling $C_v$ from the aspect of data augmentation. As shown in Alg.~\ref{alg1}, P-CSS contains two intervention procedures on visual features and reports, respectively, as follows:

\textbf{Intervention on Vision.}
As shown in Alg.~\ref{alg1}, given the visual feature of a radiograph $V$, P-CSS randomly selects one class $j$.
Then, using $j$ as a query, P-CSS retrieves the prototype of $PM_j$ from $PM$. Thus, $V$ can be divided into $K$ groups of visual features related to $K$ categories, denoted as $V = \{V_1, V_2, \dots, V_j,\dots, V_K\}$, where $V_j$ is the $j$th group.
After that, we can intervene the $j$th group of visual features as follows:
\begin{eqnarray}
\hat{V}^{(j)} = V \; \odot \;(1- PM_{j}^{T}),
\end{eqnarray}
where $\hat{V}^{(j)}$ is the intervened visual features which can be denoted as $\hat{V}^{(j)}=\{V_1, V_2, \dots, \hat{V}_j,\dots, V_K\}$. $\odot$ represents the element-wise product.



\textbf{Intervention on Report.}
To maintain the alignment of the visual feature and its report, after intervening the visual features of one prototype, we must further intervene the corresponding sentence in the report.
It is worth noting that, not all categories are described in each report, thus the following intervention procedure on the report should be conducted only when there is a describing sentence for $j$. 

For aligning the visual feature and the report, it's probably easy to imagine that we only need to pop out $y_t$ from the report. However, as aforementioned, one sentence might describe multiple categories in RRG reports. Therefore, except for solely popping out $y_t$, P-CSS must intervene the spatial features of other categories that are described in $y_t$. This implies a loop back to the Intervention on Vision, where the selected category is altered to the other described categories.

Specifically, as shown in Alg.~\ref{alg1}, P-CSS first finds the describing sentence $y_t$ from the report and obtains the labels $L_{y_t}$. Then, intervene the visual features of each item in $L_{y_t}$ until the spatial features of all described categories are intervened. After that, P-CSS pops out this sentence $y_t$.
Finally, to produce the intervened report $\hat{Y}^{(j)}_p$, P-CSS concatenates the remaining sentences in $Y_{y:l}$ as follows:
\begin{eqnarray}
\begin{aligned}
\hat{Y}^{(j)}_p &= concate([y_t]_{t=1}^{T^{'}}),\\
where \;
T^{'}&=\begin{cases}
T-1\text{,}& j $ is described$\\
T \text{,}& \text{else}
\end{cases}.
\end{aligned}
\end{eqnarray}

\subsection{Magic-Cube-like Counterfactual Report Reconstruction}
As analyzed in Section. \ref{label:cc}, the Conditional Sequential Coupling $C_s$ is a spurious confounder in the biased datasets that confuses the accuracy of models.
To break the toxic sequential relationship, we propose a Magic-Cube-like Counterfactual Report Reconstruction (Cube) strategy that is easy-but-effective and plug-and-play.
Cube randomly reconstructs the sequence of sentences, as shown in Fig.~\ref{fig:method}. 
Specifically, given the paragraph-like report $Y$, Cube first splits it into a list of sentences, denoted as:
\begin{eqnarray}
\{y_1,...,y_T\} = split(Y).
\end{eqnarray}
Then, Cube randomly disrupts the sequence of $\{y_1,...,y_T\}$, denoted as $ds(\{y_1,...,y_T\})$. Subsequently, the disrupted sentences are concatenated in sequence as the reconstructed reports $\hat{Y}_r$, formulated as:
\begin{eqnarray}
\hat{Y}_r = concate(ds(\{y_1,...,y_T\})).
\end{eqnarray}

This strategy bears a striking resemblance to disrupting a Magic-Cube, as illustrated in Fig. \ref{fig:method}, where the sequence is disrupted but the individual content in each block is unchanged, thus the Cube still remains itself in essence. Because the reconstructed reports are essentially the same as the raw reports, Cube is a model-agnostic strategy that could be arbitrarily plugged and played into other advanced models.


\begin{table*}
\centering 
\caption{Comparison of the baseline models trained on the original datasets and their results when using the P-CSS $w/ P-CSS$, the Cube $w/ Cube$, and the ensemble of them $w/ P-CSS \& Cube$. \textbf{Bold} font represents the best performance.
} 
\label{table:sota}
\makebox[\textwidth][c]{
\begin{tabular}{l|cccccc|cccc}
\hline
Methods& BLEU-1& BLEU-2& BLEU-3& BLEU-4 &METEORr &ROUGE-L & Accuracy & Precision &Recall &F1-score \\
\hline
Show-Tell  \cite{Vinyals_2015_CVPR} &  0.2177 & 0.1313 & 0.0858 & 0.0606  & 0.0935 & 0.2412 & 0.7750 & 0.1722  & 0.3697 & 0.2350   \\ 
w/ P-CSS  & 0.2193 & 0.1320 & 0.0863 & 0.0612 & 0.0963 &  0.2439 & 0.7772 & 0.1786 & 0.3822 &  0.2435\\ 
w/ Cube  & 0.1875 & 0.1144 & 0.0758 & 0.0537 &  \textbf{0.0967} &  \textbf{0.2454} &0.7767 & \textbf{0.1914} & 0.3861 & \textbf{0.2560}  \\ 
w/ P-CSS\&Cube  &  \textbf{0.2266} & \textbf{0.1358} & \textbf{0.0886} & \textbf{0.0626} & 0.0953 & 0.2434 & \textbf{0.7775} & 0.1865 & \textbf{0.3872} & 0.2517  \\ 
\hline
\hline
Top-Down  \cite{Anderson_2018_CVPR} & 0.2651& 0.1676 & 0.1141 & 0.0825 & 0.1171 & 0.2708 & 0.7705 & 0.1510 & 0.3388 & 0.2089   \\ 
w/ P-CSS & \textbf{0.2815} & \textbf{0.1772} & \textbf{0.1195} & \textbf{0.0853} & 0.1201 & 0.2724 & 0.7721 & 0.1640 & 0.3539 & 0.2242  \\ 
w/ Cube & 0.2699 & 0.2690 & 0.1139 & 0.0812 & 0.1181 & 0.2700 & 0.7746 & 0.1633 & 0.3629 & 0.2252  \\ 
w/ P-CSS\&Cube & 0.2776 & 0.1752& 0.1186 & 0.0849 & \textbf{0.1204} & \textbf{0.2728} & \textbf{0.7775} & \textbf{0.1785} & \textbf{0.3832} & \textbf{0.2436}  \\ 
\hline
\hline
Vanilla Transformer \cite{vaswani2017attention} &  0.3085 & 0.1890 & 0.1259 & 0.0896  & 0.1243 & 0.2641 & 0.7781 & 0.2302  & 0.4067 & 0.2940   \\ 
w/ P-CSS  &  0.3194 & 0.1962 & 0.1317 & 0.0948  & 0.1281 & 0.2690 & 0.7839  & 0.2491  & 0.4332 & 0.3163  \\ 
w/ Cube  &  0.3184 & 0.1941 & 0.1293 & 0.0945  & 0.1277 & 0.2653 & 0.7815 & 0.2342  & 0.4204 & 0.3008   \\ 
w/ P-CSS\&Cube  &  \textbf{0.3278} & \textbf{0.2008} & \textbf{0.1342} & \textbf{0.0964}  & \textbf{0.1301} & \textbf{0.2703} & \textbf{0.7843}  & \textbf{0.2528}  & \textbf{0.4353} & \textbf{0.3199}  \\ 
\hline
\hline
R2Gen  \cite{chen2020generating} & 0.3430 & 0.2094 & 0.1401 & 0.1003 & 0.1352 & 0.2721 & 0.7898 & 0.3017 & 0.4636 & 0.3655 \\
w/ P-CSS & 0.3591 & 0.2182 & 0.1458 & 0.1044 & 0.1396 & 0.2744 & 0.7906 & 0.3058 & 0.4668 & 0.3695 \\ 
w/ Cube & 0.3532 & 0.2148 & 0.1438 & 0.1036 & 0.1378 & 0.2727 & 0.7926 & 0.3236 & 0.4754 & 0.3851 \\ 
w/ P-CSS\&Cube & \textbf{0.3647} & \textbf{0.2214} & \textbf{0.1476} & \textbf{0.1057} & \textbf{0.1408} & \textbf{0.2743} & \textbf{0.7935} & \textbf{0.3312} & \textbf{0.4790} & \textbf{0.3916}  \\ 
\hline
\hline
R2GenCMN  \cite{chen2020generating} & 0.3179 & 0.1933 & 0.1294 & 0.0925 & 0.1291 & 0.2715 & 0.7870 & 0.2820 & 0.4508 & 0.3470 \\
w/ P-CSS &  0.3196 & 0.1949 & 0.1299 & 0.0928 & 0.1302 & 0.2692 & 0.7913 & 0.3016 & 0.4689 & 0.3671\\ 
w/ Cube & 0.3019 & 0.1859 & 0.1245 & 0.0891 & 0.1263 & 0.2693 & 0.7893 & 0.2880 & 0.4600 & 0.3542  \\ 
w/ P-CSS\&Cube  & \textbf{0.3216} & \textbf{0.1961} & \textbf{0.1307} & \textbf{0.0935} &  \textbf{0.1308} & \textbf{0.2696} & \textbf{0.7918} & \textbf{0.3063} & \textbf{0.4713} & \textbf{0.3713}  \\ 
\hline
\end{tabular}}
\end{table*}

\section{Experiments}
\subsection{Datasets}
We conduct experiments on the widely-used MIMIC-CXR dataset~\cite{johnson2019mimic}. This is the largest dataset so far, which totally contains 473,057 chest X-ray images and 206,563 reports. We adopt the annotation of R2Gen~\cite{chen2020generating} with 270,790 samples for training, 2,130 samples for validation, and 3,858 samples for testing. Each sample consists of one image and its corresponding report.

Additionally, to verify the generalization ability, we exam the models pre-tained on MIMIC-CXR dataset on the IU X-Ray dataset without finetuning. IU X-Ray is a widely used dataset that contains 7,470 Chest X-ray images and 3,955 reports, where each sample consists of two images and one report. In this paper, to meet the data format of MIMIC-CXR dataset, we select one image and its report from IU X-ray as one testing sample. The testing set contains 791 samples, which follows the annotation of R2Gen~\cite{chen2020generating}.
\subsection{Evaluation Metrics}
We evaluate our proposed approach on the widely used Neural Language Generation (NLG) metrics: BLEU \cite{papineni2002bleu}, ROUGE-L \cite{lin2004rouge} and METEOR \cite{banerjee2005meteor}. These metrics are calculated using the standard evaluation protocol~\footnote{https://github.com/tylin/coco-caption}.
Besides, Accuracy, Precision, Recall, and F1-score are taken as the metrics for Clinical Efficacy (CE) estimate. Practically, we first use the CheXbert labeler~\footnote{https://github.com/stanfordmlgroup/CheXbert} to produce the class labels for 14 diseases from the generated and ground-truth reports respectively. Then, calculate CE metrics based on the produced labels. 

\subsection{Implementation Details}
We use the ResNet-101 \cite{he2016deep} pretained on ImageNet \cite{deng2009imagenet} as the backbone CNN model to extract the visual features of radiographs. 
The dimensions of the extracted visual features are set to 2,048, and the height and width are set to 7 and 7 which are then flattened into $7\times 7 =49$. 
Moreover, models are pre-trained for 20 epochs and fine-tuned with our proposed method for 10 epochs, and both of them are supervised by cross-entropy loss and optimized by ADAM \cite{2014Adam} with a weight deacy of 5e-5. 
We set the initial learning rate to 1e-4 for the report generator and 5e-5 for ResNet-101, with a delay of 0.8 per epoch. The beam size is set to 3, and the max sequence length to be operated is 100.
The monitor to record the best performance on the validation set is the Accuracy in CE metrics.

We apply our proposed strategies to a wide range of state-of-the-art models, including Show-Tell~\cite{Vinyals_2015_CVPR}, Visual-Attention~\cite{Vinyals_2015_CVPR}, Spatial-Attention~\cite{Vinyals_2015_CVPR}, Top-Down~\cite{Anderson_2018_CVPR}, Vanilla Transformer~\cite{vaswani2017attention}, R2Gen~\cite{chen2020generating}, and R2GenCMN~\cite{ChenSSW20}. 
We implement these models and train them from scratch. 
P-CSS and Cube are applied through fine-tuning for 10 epochs on the pretrained models. During the fine-tuning stage, the learning rate of RRG models is set to 1e-5 without decay. For P-CSS, the weights of the class projection function is extracted from the pretraining results, and the parameters of CNN model are frozen during fine-tuning.

\begin{figure*}
\centering
\includegraphics[width=.97\linewidth]{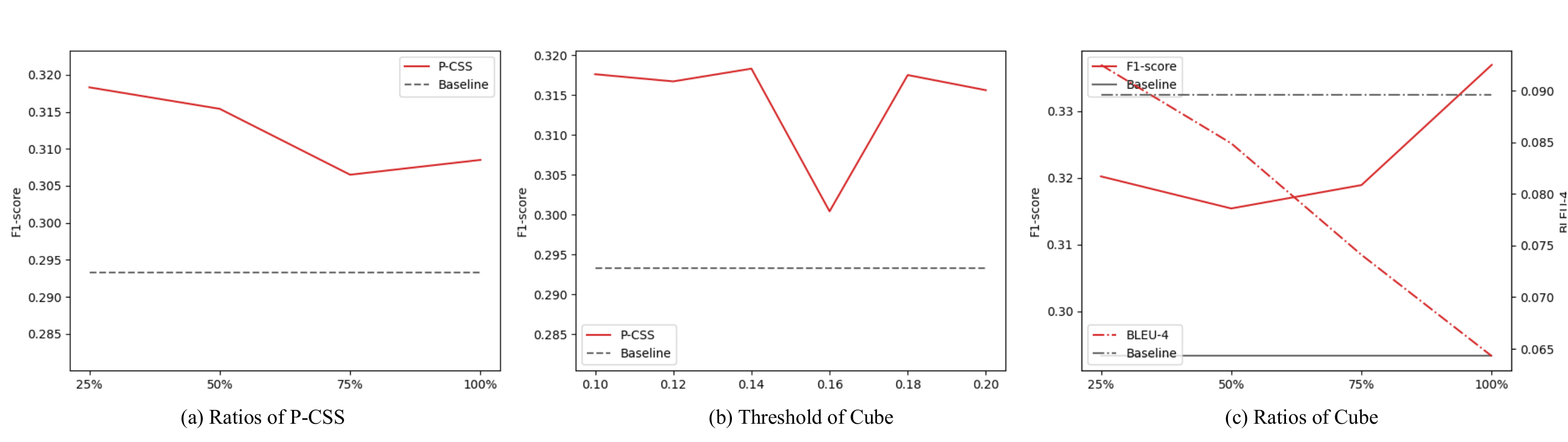}
\caption{\label{fig:hpp}{Hyper-parameter settings for P-CSS and Cube. (a) is the comparison of different ratios of P-CSS in the training set on CE (F1-score) metric. (b) is the comparison of different thresholds to establish the class-wise prototype matrix on CE metric. (c) is comparison of different ratios of Cube in the training set on the NLG (BLEU-4) and CE metrics. The red colored lines are the performance on intervened training set, and the grey colored lines are on the original set.
}}
\end{figure*}
\subsection{Effects of P-CSS}
\label{label:effectsOfPCSS}
The comparisons of applying our P-CSS strategy on baseline models are shown in Tab.~\ref{table:sota}, which are denoted as \textit{w/ P-CSS}.
It is illustrated in Tab.~\ref{table:sota} that both of the NLG metrics and the CE metrics improve a lot when RRG models apply our P-CSS strategy. 
We believe this improvement derives from our P-CSS breaks the coupling of visual features through masking the class related spatial features and the corresponding class related describing sentence.
Through counterfactual masking processing, P-CSS synthesizes new training samples that do not obey the biased data distribution of the original dataset. Therefore, RRG models can be supervised to learn more causal information from our counterfactual samples, instead of leaning to produce results confused by the spurious disease co-occurrence.
It is worth noting that, because the intervened samples disrupt the raw biased data distribution, 
It can be understood if the pre-trained models fitting the biased distribution show a decreasing performance on the out-of-distribution, especially on the NLG metrics.
However, we are delighted that the RRG models applying P-CSS do not show a decline in NLG and CE metrics in the results, which demonstrates the progressiveness of our P-CSS. 

\subsection{Effects of Cube}
\label{label:effectsOfCube}
The comparison of applying our Cube strategy to baseline models is shown in Tab.~\ref{table:sota}, which is denoted as \textit{w/ Cube}.
As illustrated in Tab.~\ref{table:sota}, the baseline models applying our Cube strategy show increasing trends, especially on the CE metrics.
This improvement indicates that our Cube has the ability to fertilize the accurate disease recognition ability of the RRG models, thus generating better reports with higher clinical trustworthiness. We attribute this improvement to the fact that our Cube breaks the coupling of sequential sentences, because of which the generation of a sentence relies more on reality with a limited impact from the previously generated sentences. Due to the fact that there are few relationships among the sentences in the report, RRG models applying our Cube can be impacted by the previous sentences fewer, thus generating a more accurate report. 
In certain baseline models, however, the decrease in NLG metrics is visible. As previously stated, the counterfactual intervention disrupts the original data distribution of the datasets, causing pre-trained models that fit the original biased distribution to fail to adapt to a comparison distribution. Because we desire better performance on CE metrics to improve the accuracy and trustworthiness of RRG models, and our Cube disrupts the spurious relationship between sentences, we suppose the decrease in NLG metrics can be understood.


\subsection{Ensemble of P-CSS and Cube}
As analyzed previously in Sec.~\ref{label:effectsOfPCSS} and ~\ref{label:effectsOfCube}, our proposed P-CSS and Cube are two model-agnostic strategies that can boost the performance of RRG models when applying them to a wide range of baseline models. The improvements on CE metrics are worthy of attention, which indicates the effectiveness of our P-CSS and Cube in breaking the spurious confounders and thus guiding RRG models to generate accurate reports. To examine the effect of the ensemble of P-CSS and Cube, we conduct the experiments on it. Practically, after training the models on the original MIMIC-CXR dataset, we apply Cube and P-CSS to the training set and fine-tune the baseline models in sequence. As shown in the Tab.~\ref{table:sota} where the ensemble of P-CSS and Cube are denoted as \textit{w/ P-CSS \& Cube}, the baseline models with the ensemble of our two strategies show increasing performances on all the comparison metrics than the models trained on the original dataset. This improvement indicates the effectiveness of the ensemble method. Then, when compared with the baseline models solely applying with P-CSS or Cube, the ensemble method shows the best performance on both NLG and CE metrics on almost all the baseline models. We attribute this performance gain to the fact that the ensemble method has the ability to combine the effectiveness of P-CSS and Cube. P-CSS and Cube respectively solve two confounders caused by the disease co-occurrence with limit interactions, thus the ensemble method can take full advantage of these two sub-methods.

\subsection{Hyper-parameter Settings}
\subsubsection{Ratios of P-CSS}~\\
The ratio of P-CSS controls the amount of training samples that are intervened by P-CSS strategy. The F1-score comparison of different ratios of P-CSS in the training set is shown in Fig.~\ref{fig:hpp} (a), where the baseline model is Vanilla Transformer and we set 25\%, 50\%, 75\%, and 100\% of the entire training samples to be intervened. The red colored line denotes the performances trained on intervened dataset while the grey colored line is the one trained on the original dataset. Overall, applying all ratios of P-CSS increases the F1-score of the baseline model, which indicates the model generates more accurate reports. Then, it can be observed that the model applying with 25\% of P-CSS shows the best CE performance. As the proportion increases, the F1-score decreases but still exceeds the model trained on the original dataset. Therefore, it is advised to set the ratio of P-CSS to 25\%.
\subsubsection{Thresholds of P-CSS}~\\
The threshold $\tau$ of P-CSS controls the establishment of a class-wise prototype matrix, where a higher $\tau$ leads to fewer spatial features to be intervened.
The F1-score comparison results of different values of $\tau$ on the CE performance are shown in Fig.~\ref{fig:hpp} (b), where we set $\tau$ to 0.10, 0.12, 0.14, 0.16, 0.18, and 0.20. The color settings of lines are the same as the ratios of P-CSS. From Fig.~\ref{fig:hpp} (b) we can find that the baseline model applying all the settings of $\tau$ exceeds the baseline model trained on the original dataset. Then, it can be observed that when setting $\tau$ into 0.14, the model obtains the best performance. Thus, we advise to set $\tau$ to 0.14.

\begin{figure*}
\centering
\includegraphics[width=.71\linewidth]{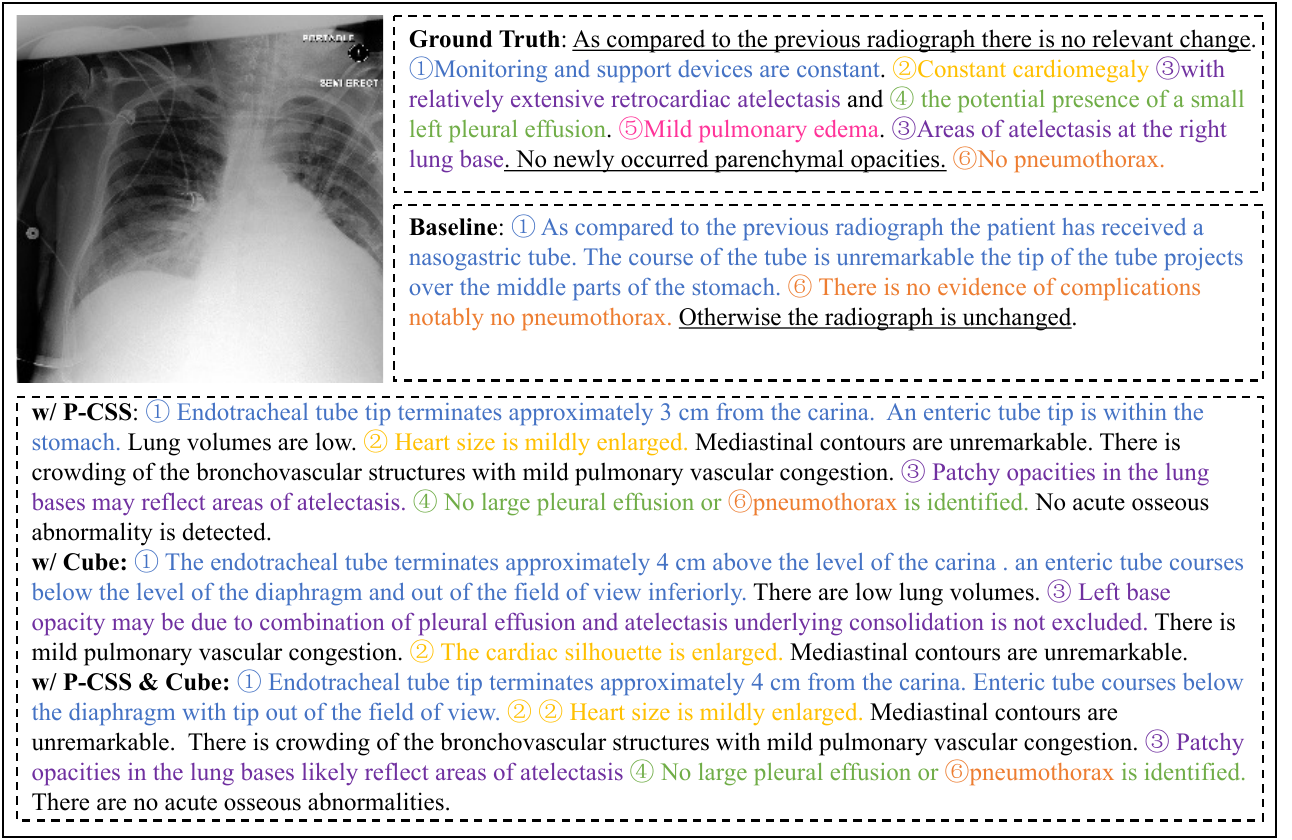}
\caption{\label{fig:vis1}{Visualization of the comparison of ground truth reports, the reports generated by the baseline model and the model applying our method on MIMIC-CXR dataset. The serial numbers with various colors are sentences describing the abnormal findings. The underlined sentences contain information that does not present in the radiographs.
}}
\end{figure*}
\subsubsection{Ratios of Cube}~\\
The ratio of Cube regulates how many training samples the Cube approach intervenes with. In Fig.~\ref{fig:hpp} (c), where we set 25\%, 50\%, 75\%, and 100\% of the complete training samples to be intervened. The difference ratios of Cube in the training set are compared on both NLG (BLEU-4) and CE metrics which are respectively denoted as solid and dashed lines. The color settings of lines are the same as the ratios of P-CSS. It can be observed that the model with higher Cube intervening portions gains higher CE metrics, which indicates that the model generates more accurate reports. However, we find that the higher ratios lead to lower NLG metrics, which can be explained as aforementioned in Sec.~\ref{label:effectsOfCube}.
To comprehensively consider the performances of the model in terms of both CE and NLG metrics, we suggest to set the ratios of Cube to 25\%.

\begin{table}
\centering 
\caption{Comparison of different parameters that are trained on the MIMIC-CXR dataset on IU X-Ray dataset without fine-tuning to exam the generalization performances. } 
\label{table:generalization}
\begin{tabular}{l|cccc}
\hline
Pre-trained & Accuracy & Precision & Recall & F1-score\\
\hline
Baseline & 0.9271 & 0.5302 & 0.5086 & 0.5192\\
w/P-CSS & 0.9322 & 0.5318 & 0.5422 & 0.5380\\
w/Cube & 0.9320 & 0.5367 & 0.5420 & 0.5393\\
w/P-CSS \& Cube & \textbf{0.9329} & \textbf{0.5351} & \textbf{0.5494} & \textbf{0.5421}\\
\hline
\end{tabular}
\end{table}

\subsection{Generalization Ability}
Training on biased datasets will force deep learning models to fit the biased distribution for lower objective loss.
Therefore, when modifying the biased distribution, it is unquestionable that RRG models will show a performance decline on the original datasets.
This decline does not, however, imply that the intervention methods are insufficient or ineffective for the objective or goal.
In contrast, if RRG models have learned to equip themselves with causal recognition abilities about diseases, their learned parameters should have better generalization abilities.

In order to comprehensively study the generalization ability of P-CSS and Cube in mining causal disease information, we directly test the Vanilla Transformer and parameters trained on MIMIC without fine-tuning on IU X-Ray dataset. It is worth noting that due to the fact that the report formats of IU X-Ray dataset and MIMIC-CXR dataset are different, thus it is of little value to exam the NLG metrics.
The results of CE metrics are shown in Tab.~\ref{table:generalization}. It can be found that the baseline model that uses our P-CSS or Cube outperforms the baseline model trained on the original dataset. Additionally, applying the ensemble of P-CSS and Cube achieves the best performance among the contrasts, which is different from the testing performance on MIMIC-CXR dataset. Compared to the previous works, our model-agnostic P-CSS and Cube somewhat mitigate the bias in the data, making them plug-and-play tools for future research.

\subsection{Visualization}
We also visualize the report generated by the baseline Vanilla model and its combinations with our method, where the results are shown in Fig. \ref{fig:vis1}. First, from the comparison of colored sentences with serial numbers, it is obvious that models applying our method could effectively identify more disease information and generate more accurate reports. Additionally, compared to the underlined sentences that describe the disease progression between multiple examinations, the baseline models generate more disease progression sentences that can not be identified based on a single radiograph. This phenomenon means that the ground truth reports include some amounts of spurious and confusing information that could not be found in the corresponding radiographs, hurting the models' performance. Fortunately, models applied with our P-CSS and Cube show much fewer underlined sentences. We assume that this performance improvement could give credit to the fact that our method helps the models focus on the critical visual features and reduce the dependence between sentences, instead of generating reports based on the co-occurrence of diseases. 
Besides, it can be seen from Fig. \ref{fig:vis1} that the models applying our P-CSS and Cube generate more comprehensive descriptions compared to the ground-truth report and the baseline results. For example, $w/P-CSS$ and $w/ P-CSS \& Cube$ describe the information of acute and mediastinal counters.

\section{Conclusion}
Radiology Report Generation (RRG) is noteworthy as a vision-and-language connection of biomedical fields and previous works have made significant progress in generating paragraphs with high readability as reports. 
However, these works neglected the independence between diseases—a specific property of RRG, leading the models be confused by the co-occurrence of diseases brought on by the biased data distribution, thus decreasing the accuracy of the generated reports.
In this paper, we first analyze the co-occurrence relationships between diseases on the biased distribution from a causal view and then a novel model-agnostic counterfactual augmentation method is developed for radiology report generation tasks. 
By integrating the designed prototype-based counterfactual sample synthesis (P-CSS) and the magic-cube-like counterfactual report reconstruction (Cube) strategies, the accuracy and generalization of RRG models are improved. Experimental results on MIMIC-CXR and IU X-Ray datasets demonstrate the effectiveness of our proposed method. 
To our knowledge, this is the first work to discuss the disease co-occurrence in the RRG task from a causal perspective. Based on the exploration of this work, we desire to explore the more fine-grained causal relationships between variables in future works in order to improve the accuracy and trustworthiness of RRG models.
\bibliographystyle{ACM-Reference-Format}
\bibliography{sample-base}

\end{document}